\documentclass{article}

\usepackage[preprint]{nips_2018}

\usepackage[utf8]{inputenc} 
\usepackage[T1]{fontenc}    
\usepackage{hyperref}       
\usepackage{url}            
\usepackage{booktabs}       
\usepackage{amsfonts}       
\usepackage{nicefrac}       
\usepackage{microtype}      

\usepackage{graphicx}
\usepackage{subfigure}
\usepackage{xcolor}
\usepackage{amsmath}
\usepackage{multirow}
\usepackage{enumitem}
\usepackage{tabularx}
\usepackage{makecell}

\usepackage{bm}
\usepackage{bbm}
\usepackage{color}
\newcommand{\itmvae}{iTM-VAE\xspace}
\newcommand{\itmvaeprod}{iTM-VAE-Prod\xspace}
\newcommand{\itmvaefull}{iTM-VAE-HP\xspace}
\newcommand{\itmvaehie}{HiTM-VAE\xspace}

\newcommand{\reals}{\mathbb{R}}

\newcommand{\expect}{\mathbb{E}}

\newcommand{\loss}{\mathcal{L}}
\newcommand{\kumar}{Kumaraswamy\xspace}
\usepackage{float}
\usepackage{multicol}
\floatstyle{plain}
\newfloat{mixedcol}{thp}{lop}

\usepackage{xspace}

\setcitestyle{numbers,square,comma}

\title{Nonparametric Topic Modeling with Neural Inference}

\author{
  Xuefei Ning\\
  Tsinghua University\\
  \texttt{foxdoraame@gmail.com} \\
   \And
   Yin Zheng \\
   Tencent AI Lab \\
   \texttt{yzheng3xg@gmail.com} \\
   \AND
   Zhuxi Jiang \\
   Momenta \\
   \texttt{zjiang9310@gmail.com} \\
   \And
   Yu Wang \\
   Tsinghua University \\
   \texttt{yu-wang@tsinghua.edu.cn} \\
   \And
   Huazhong Yang \\
   Tsinghua University \\
   \texttt{yanghz@tsinghua.edu.cn } \\
   \And
   Junzhou Huang \\
   Tencent AI Lab \\
   \texttt{joehhuang@tencent.com} \\
}

\begin{document}

\maketitle

\begin{abstract}
This work focuses on combining nonparametric topic models with Auto-Encoding Variational Bayes (AEVB).
Specifically, we first propose \itmvae, where the topics are treated as trainable parameters and
the document-specific topic proportions are obtained by a stick-breaking construction. The inference of \itmvae
is modeled by neural networks such that it can be computed in a simple feed-forward manner. 
We also describe how to introduce a 
hyper-prior into \itmvae so as to
model the uncertainty of the prior parameter. Actually, the hyper-prior technique
is quite general and we show that it can be applied to other AEVB based models to alleviate
the {\it collapse-to-prior} problem elegantly. Moreover, we also propose \itmvaehie, where
the document-specific topic distributions are generated in a hierarchical manner. \itmvaehie is even
more flexible and can generate topic distributions with better variability. Experimental results on 20News and Reuters RCV1-V2
datasets show that the proposed models outperform the state-of-the-art baselines significantly.
The advantages of the hyper-prior technique and the hierarchical model construction are also confirmed by experiments.

\end{abstract}

\section{Introduction}
\label{sec:intro}

Probabilistic topic models focus on discovering the abstract ``topics'' that occur in a collection of 
documents, and represent a document as a weighted mixture of the discovered topics.
Classical topic models~\citep{blei2003lda}
have achieved success in a range of applications~\citep{wei2006lda,blei2003lda,rasiwasia2013latent,rogers2005latent}. 
A major challenge of topic models is that 
the inference of the distribution over topics does not have a closed-form solution and
must be approximated, using either MCMC sampling or variational inference.
When some small changes are made on the model, we need to re-derive the inference algorithm.
In contrast, black-box inference methods~\citep{ranganath2014black,mnih2014neural,kingma2013auto,rezende2014stochastic} require only limited model-specific analysis and
can be flexibly applied to new models.



Among all the black-box inference methods, Auto-Encoding Variational Bayes (AEVB)~\citep{kingma2013auto,rezende2014stochastic}
is a promising one for topic models. AEVB contains an inference network that can map a document directly to 
a variational posterior without the need for further local variational updates on test data, and the Stochastic Gradient Variational 
Bayes (SGVB) estimator allows efficient approximate inference for a broad class of posteriors, which makes
 topic models more flexible.
Hence, an increasing number of models are proposed recently to combine topic models with AEVB,
such as \citep{miao2016nvdm,srivastava2017avitm,card2017neural,miao2017discovering}.

Although these AEVB based topic models achieve promising performance,
the number of topics, which is important to the performance of these models, has to be specified manually with model selection methods. 
Nonparametric models, however, have the ability of adapting the topic number to data.
For example, \citet{Teh2006hdp} proposed Hierarchical Dirichlet Process (HDP),
which models each document with a Dirichlet Process (DP) and all DPs for the documents in a corpus share a base distribution that is itself sampled from a 
DP.
HDP has potentially an {\it infinite} number of topics 
and allows the number to grow as more documents are observed. 
It is appealing that the nonparametric topic models can also be equipped with AEVB techniques to enjoy the
benefit brought by neural black-box inference. 
We make progress 
on this problem by proposing an {\it infinite Topic Model with Variational Auto-Encoders} (\itmvae), which is 
a nonparametric topic model with AEVB. 



For nonparametric topic models with stick breaking prior~\citep{sethuraman1994constructive},
the concentration parameter $\alpha$ plays an important role in deciding the
growth of topic numbers\footnote{\label{note:gem_alpha}Please refer to 
Section~\ref{sec:itmvae_gen} 
for more details about the concentration parameter.}. The larger the 
$\alpha$ is, the more topics the model tends to discover. Hence, people can place a hyper-prior~\citep{bernardo2001bayesian}
over $\alpha$ such that the model can adapt it to data~\citep{escobar1995bayesian,Teh2006hdp,blei2006variational}. 
Moreover, the AEVB framework suffers from the problem that the latent representation tends
to collapse to the prior~\citep{bowman2015generating,sonderby2016ladder,chen2016variational}, which means, the prior parameter $\alpha$ will control the number of discovered topics tightly in our case, especially when the decoder is strong. 
Common heuristic tricks to alleviate this issue are 1) KL-annealing~\citep{sonderby2016ladder} and 2) decoder
regularizing~\citep{bowman2015generating}. 
Introducing a hyper-prior
into the AEVB framework is nontrivial and not well-done in the community.  
In this paper, we show that 
introducing a hyper-prior can increase the adaptive capability of the model, and also alleviate the {\it collapse-to-prior} issue in the training process.\footnote{The hyper-prior technique can also alleviate the {\it collapse-to-prior} issue in other scenarios, an example is demonstrated in Appendix 1.2.}

To further increase the flexibility of \itmvae, we propose \itmvaehie, which model the document-specific
topic distribution in a hierarchical manner. This hierarchical construction can help to generate topic distributions with better variability, which is more suitable in handling heterogeneous documents.


The main contributions of the paper are:
\begin{itemize}
\item We propose \itmvae and \itmvaeprod, which are two novel nonparametric topic models equipped with AEVB, and outperform the state-of-the-art models on the benchmarks.
\item We propose \itmvaefull, in which a hyper-prior helps the model to adapt the prior parameter to data.
We also show that this technique 
can help other AEVB-based models to alleviate the {\it collapse-to-prior} problem elegantly.
\item We propose \itmvaehie, which is a hierarchical extension of \itmvae. This construction and its corresponding AEVB-based inference method can help the model to learn more topics and produce topic proportions with higher variability and sparsity.
\end{itemize}

\section{Related Work}
\label{sec:related}

Topic models have been studied extensively in a variety of applications such as document modeling,
information retrieval, computer vision and bioinformatics~\citep{blei2012probabilistic,blei2003lda,wei2006lda,putthividhy2010topic,rasiwasia2013latent,rogers2005latent}. 
Recently, with the impressive success of deep learning, the proposed neural topic models~\citep{hinton2009replicated,larochelle2012DNADE,mnih2014neural}
achieve encouraging performance in document modeling tasks.
Although these models achieve competitive performance, 
they do not explicitly model the generative story of documents, hence are less explainable.

Several recent work proposed to model the generative procedure explicitly, and the inference of the topic distributions in these models is computed by deep neural networks, which makes these models explainable, powerful and easily extendable. 
For example, \citet{srivastava2017avitm} 
proposed AVITM, which embeds the original LDA~\citep{blei2003lda} formulation with AEVB. 
By utilizing Laplacian approximation for the Dirichlet distribution,  AVITM can be optimized by the SGVB estimator efficiently. 
AVITM achieves the state-of-the-art performance on the 
topic coherence metric~\citep{lau2014machine},
which indicates the topics learned match closely to human judgment.

Nonparametric topic models~\citep{Teh2006hdp,kim2011doubly,archambeau2015latent,lim2016nonparametric}, 
potentially have infinite topic capacity and 
can adapt the topic number to data. 
\citet{nalisnick2016stick} proposed Stick-Breaking VAE (SB-VAE), which is a Bayesian nonparametric version of traditional VAE with a stochastic dimensionality.
\itmvae differs with SB-VAE in 3 aspects: 1) \itmvae is a kind of {\it topic model} for discrete text data. 2) A hyper-prior is introduced into the AEVB framwork to increase the adaptive capability. 3) A hierarchical extension of \itmvae is proposed to further increase the flexibility. 
\citet{miao2017discovering} proposed GSM, GSB, RSB and RSB-TF to model
documents. 
RSB-TF 
uses a heuristic indicator to guide the growth of the topic numbers, and can adapt the topic number to data.

\section{The \itmvae Model}
\label{sec:model}
In this section, 
we describe the generative and inference procedure of \itmvae and \itmvaeprod in Section~\ref{sec:itmvae_gen} and Section~\ref{sec:itmvae_inf}. Then, Section~\ref{sec:itmvae_prior} describes the hyper-prior extension \itmvaefull.


\subsection{The Generative Procedure of \itmvae}
\label{sec:itmvae_gen}



Suppose the atom weights $\boldsymbol{\pi}=\{\pi_k\}_{k=1}^\infty$ are drawn from a GEM distribution~\cite{murphy2012machine},
i.e. $\boldsymbol{\pi}\sim \text{GEM}(\alpha)$, where the GEM distribution is defined as:
\begin{equation}
\nu_k\sim \text{Beta}(1,\alpha) \qquad
\pi_k = \nu_k \prod_{l=1}^{k-1}(1-\nu_l)=\nu_k(1-\sum_{l=1}^{k-1}\pi_l)\label{eqn:gem_pi}.
\end{equation}
Let $\boldsymbol{\theta}_k = \sigma(\boldsymbol{\phi}_k)$ denotes the $k$th topic, which is a multinomial distribution over vocabulary,
$\boldsymbol{\phi}_k\in \reals^V$ is the parameter of $\boldsymbol{\theta}_k$, $\sigma(\cdot)$ is the softmax function and $V$ is 
the vocabulary size. In \itmvae, there are unlimited number of topics and we denote
$\Theta=\{\boldsymbol{\theta}_k\}_{k=1}^\infty$ and $\Phi = \{\boldsymbol{\phi}_k\}_{k=1}^\infty$ 
as the collections of these countably infinite topics and the corresponding parameters. 
The generation of a document $\boldsymbol{x}^{(j)}=\boldsymbol{w}^{(j)}_{1:N^{(j)}}$ by \itmvae can then be
mathematically described as:

\begin{itemize}
\setlength\itemsep{0.1em}

\item Get the document-specific $G^{(j)}(\boldsymbol{\theta};\boldsymbol{\pi}^{(j)},\Theta)=\sum_{k=1}^{\infty}\pi_k^{(j)}\delta_{\boldsymbol{\theta}_k}(\boldsymbol{\theta})$, where $\boldsymbol{\pi}^{(j)}\sim \text{GEM}(\alpha)$

\item For each word $w_n$ in $\boldsymbol{x}^{(j)}$: 1) draw a topic $\hat{\boldsymbol{\theta}}_n \sim G^{(j)}(\boldsymbol{\theta};\boldsymbol{\pi}^{(j)},\Theta)$;
~2) $w_n \sim \text{Cat}(\hat{\boldsymbol{\theta}}_n)$ 
\end{itemize}
where $\alpha$ is the concentration parameter, $\text{Cat}(\hat{\boldsymbol{\theta}}_i)$ is a categorical distribution parameterized by $\hat{\boldsymbol{\theta}}_i$,
and $\delta_{\boldsymbol{\theta}_k}(\boldsymbol{\theta})$ is a discrete dirac function, which 
equals to $1$ when $\boldsymbol{\theta}=\boldsymbol{\theta}_k$ and $0$ otherwise. In the following, we remove the superscript of 
$j$ for simplicity.


Thus, the joint probability of ${\bf w}_{1:N} = \{w_n\}_{n=1}^{N}$, $\hat{\boldsymbol{\theta}}_{1:N}=\{\hat{\boldsymbol{\theta}_n}\}_{n=1}^N$ and $\boldsymbol{\pi}$ can be written as:
\begin{equation}
p({\bf w}_{1:N}, \boldsymbol{\pi},\hat{\boldsymbol{\theta}}_{1:N}|\alpha,\Theta) = p(\boldsymbol{\pi}|\alpha)\prod_{n=1}^N p(w_n|\hat{\boldsymbol{\theta}}_n)p(\hat{\boldsymbol{\theta}}_n|\boldsymbol{\pi},\Theta)
\label{eqn:prob_gen}
\end{equation}
where $p(\boldsymbol{\pi}|\alpha)=\text{GEM}(\alpha)$, $p(\boldsymbol{\theta}|\boldsymbol{\pi},\Theta)=G(\boldsymbol{\theta};\boldsymbol{\pi},\Theta)$
and $p(w|\boldsymbol{\theta})=\text{Cat}(\boldsymbol{\theta})$. 

Similar to \citep{srivastava2017avitm}, we collapse the variable
$\hat{\boldsymbol{\theta}}_{1:N}$ and rewrite Equation~\ref{eqn:prob_gen} as:
\begin{equation}
p({\bf w}_{1:N}, \boldsymbol{\pi}|\alpha,\Theta) = p(\boldsymbol{\pi}|\alpha)\prod_{n=1}^N p(w_n|\boldsymbol{\pi},\Theta)
\label{eqn:collapse_prob}
\end{equation}
where $p(w_n|\boldsymbol{\pi},\Theta) = \text{Cat}(\bar{\boldsymbol{\theta}})$ and
$\bar{\boldsymbol{\theta}} = \sum_{k=1}^\infty \pi_k \boldsymbol{\theta}_k$.

In Equation~\ref{eqn:collapse_prob}, $\bar{\boldsymbol{\theta}}$ is a mixture of multinomials. This formulation 
cannot make any predictions that are sharper than the distributions being mixed~\citep{hinton2009replicated},
which may result in some topics that are of poor quality. 
Replacing the mixture of multinomials with a weighted product of experts is one method to make sharper predictions~\citep{hinton2006training,srivastava2017avitm}. Hence, a products-of-experts version of \itmvae
(i.e. \itmvaeprod) can be obtained by simply computing $\hat{\boldsymbol{\theta}}$ for each document as 
$\hat{\boldsymbol{\theta}} = \sigma(\sum_{k=1}^\infty \pi_k \boldsymbol{\phi}_k)$. 

\subsection{The Inference Procedure of \itmvae}
\label{sec:itmvae_inf}
In this section, we describe the inference procedure of \itmvae, i.e. how to draw 
${\boldsymbol{\pi}} $ given a document ${\bf w}_{1:N}$.
To elaborate, suppose $\boldsymbol{\nu}=[\nu_1,\nu_2,\ldots,\nu_{K-1}]$ is a $K-1$ dimensional vector, where
${\nu_k}$ is a random variable sampled from a \kumar distribution $\kappa(\nu;a_k,b_k)$
parameterized by $a_k$ and $b_k$~\citep{kumaraswamy1980generalized,nalisnick2016stick}, \itmvae models the joint distribution $q_{\psi}(\boldsymbol{\nu}|{\bf w}_{1:N})$ as:
\footnote{Ideally, Beta distribution
is the most suitable probability candidate, since \itmvae assumes $\boldsymbol{\pi}$ is drawn
from a GEM distribution in the generative procedure. However, as Beta does not satisfy the {\it differentiable, non-centered parameterization} (DNCP)~\citep{kingma2014efficient} requirement of SGVB~\citep{kingma2013auto},
we use the \kumar distribution.}
\begin{eqnarray}
[a_1,\ldots,a_{K-1};b_1,\ldots,b_{K-1}]&=&g({\bf w}_{1:N};\psi)\label{eqn:kumar_parameter}\\
q_{\psi}(\boldsymbol{\nu}|{\bf w}_{1:N}) &=& \prod_{k=1}^{K-1} \kappa(\nu_k;a_k,b_k)
\label{eqn:q_nu}
\end{eqnarray}
where $g({\bf w}_{1:N};\psi)$ is a neural network with parameters $\psi$.
Then, ${\boldsymbol{\pi}} = \{\pi_k\}_{k=1}^K $ can be drawn by:
\begin{eqnarray}
\boldsymbol{\nu}&\sim& q_{\psi}(\boldsymbol{\nu}|{\bf w}_{1:N})\\
{\boldsymbol{\pi}} &=& ({\pi}_1,{\pi}_2,\ldots,{\pi}_{K-1},{\pi}_K) =
(\nu_1,\nu_2(1-\nu_1),\ldots,\nu_{k-1}\prod_{n=1}^{K-2}(1-\nu_n),\prod_{n=1}^{K-1}(1-\nu_n))
\label{eqn:inf_pi}
\end{eqnarray}

In the above procedure, we truncate the infinite sequence of mixture weights $\boldsymbol{\pi}=\{\pi_k\}_{k=1}^\infty$ by $K$ elements,
and $\nu_{K}$ is always set to $1$ to ensure $\sum_{k=1}^K {\pi}_k=1$. Notably, as is  discussed in \citep{blei2006variational},
the truncation of variational posterior does {\it not} indicate that we are using a finite dimensional prior, since we {\it never} 
truncate the GEM prior. 
Hence, \itmvae still has the ability to model the uncertainty of the number of topics and adapt it to data~\citep{nalisnick2016stick}.

\itmvae can be optimized by maximizing the Evidence Lower Bound (ELBO):
\begin{eqnarray}
\loss({\bf w}_{1:N}|\Phi,\psi) = \expect_{q_{\psi}(\boldsymbol{\nu}|{\bf w}_{1:N})}\left[
\log p({\bf w}_{1:N}| \boldsymbol{\pi},\Phi)\right] 
- \text{KL}\left(q_{\psi}(\boldsymbol{\nu}|{\bf w}_{1:N})||p({\boldsymbol{\nu}|\alpha})\right)
\label{eqn:itmvae_elbo}
\end{eqnarray}
where $p({\boldsymbol{\nu}|\alpha})$ is the product of
 $K-1$ $\text{Beta}(1,\alpha)$ probabilistic density functions. 
The details of the optimization can be found in Appendix 1.3.

\subsection{Modeling the Uncertainty of Prior Parameter}
\label{sec:itmvae_prior}

In the generative procedure, the concentration parameter $\alpha$ of $\text{GEM}(\alpha)$ can 
have significant impact on the growth of number of topics. The larger the $\alpha$ is, the more ``breaks" it will 
create, and consequently, more topics will be used. Hence, it is generally reasonable to consider placing 
a hyper-prior on $\alpha$ to model its uncertainty.\cite{escobar1995bayesian,blei2006variational,Teh2006hdp}. 
For example, \citet{escobar1995bayesian} placed a Gamma hyper-prior on $\alpha$ for 
the urn-based samplers and implemented the corresponding Gibbs updates with auxiliary variable methods.
\citet{blei2006variational} also placed a Gamma prior on $\alpha$ and derived a closed-form update for the 
variational parameters. Different with previous work, we introduce the hyper-prior into the AEVB framework and 
propose to optimize the model by stochastic gradient decent (SGD) methods.

Concretely, since the Gamma distribution is conjugate to $\text{Beta}(1,\alpha)$, we place
a $\text{Gamma}(s_1,s_2)$ prior on $\alpha$. Then the ELBO of \itmvaefull can be written as:
\begin{eqnarray}
\loss({\bf w}_{1:N}|\Phi,\psi) &=& \expect_{q_{\psi}(\boldsymbol{\nu}|{\bf w}_{1:N})}\left[\log p({\bf w}_{1:N}| \boldsymbol{\pi}, \Phi)\right] + \mathbb{E}_{q_{\psi}(\boldsymbol {\nu}|{\bf w}_{1:N})q(\alpha|\gamma_1
	,\gamma_2)} [\log p(\boldsymbol{\nu}|\alpha)] \nonumber\\
& & - \mathbb{E}_{q_{\psi}(\boldsymbol{\nu}|{\bf w}_{1:N})}[\log q_{\psi}(\boldsymbol{\nu}|{\bf w}_{1:N})] - \text{KL}(q(\alpha|\gamma_1,\gamma_2)||p(\alpha|s_1,s_2))
\label{eqn:itmvae_elbo_fb}
\end{eqnarray}
where $p(\alpha|s_1,s_2)=\text{Gamma}(s_1,s_2)$, $p(v_k|\alpha)=\text{Beta}(1,\alpha)$, $q(\alpha|\gamma_1,\gamma_2)$ is
the corpus-level variational posterior for $\alpha$. 
The derivation for Equation~\ref{eqn:itmvae_elbo_fb} can be found in Appendix 1.4. 
In our experiments, we find \itmvaeprod
always performs better than \itmvae, therefore we only report the performance of \itmvaeprod with hyper-prior, and refer this variant as
\itmvaefull.
Actually, as discussed in Section~\ref{sec:intro}, the hyper-prior technique can also be applied to other AEVB based models
 to alleviate the {\it collapse-to-prior} problem. 
In Appendix 1.2, we show that by introducing a hyper-prior to SB-VAE, more latent units can be activated and the model achieves
better performance.








\section{Hierarchical \itmvae}
\label{sec:itmvae_hie}

In this section, we describe the generative and inference procedures 
of \itmvaehie in Section~\ref{sec:itmvaehie_gen} and Section~\ref{sec:itmvaehie_inf}. 
The relationship between \itmvae and \itmvaehie is discussed in Section~\ref{sec:discuss}


\subsection{The Generative Procedure of \itmvaehie}
\label{sec:itmvaehie_gen}
The generation of a document by \itmvaehie is described as follows:
\begin{itemize}
    \item Get the corpus-level base distribution $G^{(0)}$: $\boldsymbol{\beta} \sim \mbox{GEM}(\gamma); G^{(0)}(\boldsymbol{\theta}; \boldsymbol{\beta}, \Theta) = \sum_{i=1}^\infty \beta_i \delta_{\boldsymbol{\theta}_i}(\boldsymbol{\theta})$
    \item For each document $\boldsymbol{x}^{(j)}=\boldsymbol{w}^{(j)}_{1:N^{(j)}}$ in the corpus: 
    \begin{itemize}
        \item Draw the document-level stick breaking weights  $\boldsymbol{\pi}^{(j)} \sim \mbox{GEM}(\alpha)$
        \item Draw document-level atoms $\boldsymbol{\zeta}_{k}^{(j)} \sim G_0$, $k=1,\cdots,\infty$; Then we get a document-specific distribution $G^{(j)}(\boldsymbol{\theta}; \boldsymbol{\pi}^{(j)}, \{\boldsymbol{\zeta}_{k}^{(j)}\}_{k=1}^{\infty}, \Theta) = \sum_{k=1}^\infty \pi_{k}^{(j)} \delta_{\boldsymbol{\zeta}_{k}^{(j)}}(\boldsymbol{\theta})$
        \item For each word $w_n$ in the document: 1) draw a topic $\hat{\boldsymbol{\theta}}_n \sim G^{(j)}$; 2) $w_n \sim \mbox{Cat}(\hat{\boldsymbol{\theta}}_n)$
    \end{itemize}
\end{itemize}

To sample the document-level atoms $\boldsymbol{\zeta}^{(j)} = \{\boldsymbol{\zeta}_{k}^{(j)}\}_{k=1}^\infty$
, a series of indicator variables ${\bold c^{(j)}} = \{c_{k}^{(j)}\}_{k=1}^\infty$ 
are drawn i.i.d: $c_{k}^{(j)} \sim \mbox{Cat}(\boldsymbol{\beta})$. Then, the document-level atoms  
are $\boldsymbol{\zeta}_{k}^{(j)} = \boldsymbol{\theta}_{c_{k}^{(j)}}$.

Let $D$ and $N^{(j)}$ denote the size of the dataset and the number of word in each document $\boldsymbol{x}^{(j)}$, respectively. After collapse the  per-word assignment random variables $\{\{\hat{\boldsymbol{\theta}}_{n}^{(j)}\}_{n=1}^{N^{(j)}}\}_{j=1}^D$, the joint probability of the corpus-level atom weights $\boldsymbol{\beta}$, documents $\mathcal{X}=\{\boldsymbol{x}^{(j)}\}_{j=1}^{D}$, the stick breaking weights $\Pi=\{\boldsymbol{\pi}^{(j)}\}_{j=1}^D$
and the indicator variables $\mathcal{C}=\{\boldsymbol{c}^{(j)}\}_{j=1}^D$ can be written as:
\begin{eqnarray}
p(\boldsymbol{\beta}, 
\mathcal{X}, \Pi, \mathcal{C}|\gamma,\alpha,\Theta) = p(\boldsymbol{\beta}|\gamma) \prod_{j=1}^D  p(\boldsymbol{\pi}^{(j)}|\alpha) p(\boldsymbol{c}^{(j)} | \boldsymbol{\beta}) p(\boldsymbol{x}^{(j)} | \boldsymbol{\pi}^{(j)}, \boldsymbol{c}^{(j)}, \Theta)
\label{eqn:hie_prob_gen}
\end{eqnarray}

where $p(\boldsymbol{\beta}|\gamma) = \text{GEM}(\gamma)$, $p(\boldsymbol{\pi}^{(j)}|\alpha)=\text{GEM}(\alpha)$, $p(\boldsymbol{c}^{(j)}|\boldsymbol{\beta}) = \text{Cat}(\boldsymbol{\beta})$, 
$p(x^{(j)}|\boldsymbol{\pi}^{(j)}, \boldsymbol{c}^{(j)}, \Theta) = \prod_{n=1}^{N^{(j)}} p(w_{n}^{(j)}|\boldsymbol{\pi}^{(j)}, \boldsymbol{c}^{(j)}, \Theta) = \prod_{n=1}^{N^{(j)}} \text{Cat}(w_n^{(j)}|\bar{\boldsymbol{\theta}}^{(j)}) = \prod_{l=1}^{N^{(j)}} \text{Cat}(w_n^{(j)}|\sum_{k=1}^\infty \pi_{k}^{(j)} \boldsymbol{\theta}_{c_{k}^{(j)}})$.



\subsection{The Inference Procedure of \itmvaehie}
\label{sec:itmvaehie_inf}
Setting the truncation level of the corpus-level and document-level GEM to $T$ and $K$, \itmvaehie models the per-document posterior $q(\boldsymbol{\nu}, {\bold c}|{\bf w}_{1:N})$ for every document ${\bf w}_{1:N}$ as:

\begin{eqnarray}
[a_1,\ldots,a_{K-1};b_1,\ldots,b_{K-1};\boldsymbol{\varphi}_1,\ldots,\boldsymbol{\varphi}_{K}]&=&g({\bf w}_{1:N};\psi)\label{eqn:kumar_parameter}\\
q(\boldsymbol{\nu}, {\bold c}|{\bf w}_{1:N}) &=& q_{\psi}(\boldsymbol{\nu}|{\bf w}_{1:N}) q_{\psi}(\boldsymbol{c}|{\bf w}_{1:N})\\
q_{\psi}(\boldsymbol{\nu}|{\bf w}_{1:N}) = \prod_{k=1}^{K-1} \kappa(\nu_k;a_k,b_k); \quad q_{\psi}(\boldsymbol{c}|{\bf w}_{1:N}) &=& \prod_{k=1}^{K} \mbox{Cat}(c_k; \boldsymbol{\varphi}_k)
\label{eqn:hie_q_nu}
\end{eqnarray}
where $g({\bf w}_{1:N};\psi)$ is a neural network with parameters $\psi$, and $\boldsymbol{\varphi}_k = \{\varphi_{ki}\}_{i=1}^T$ are the multinomial variational parameters for each document-level indicator variable $c_k$. Then, ${\boldsymbol{\pi}} = \{\pi_k\}_{k=1}^K $ can be constructed by the stick breaking process using $\boldsymbol{\nu}$.

As we shown in Section~\ref{sec:itmvaehie_gen}, the generation of the corpus-level atom weights $\boldsymbol{\beta}$ is as follows:
\begin{eqnarray}
\beta'_i \sim \text{Beta}(1, \gamma); \quad \beta_i = \beta'_i \prod_{l=1}^{i-1}(1-\beta'_l)
\end{eqnarray}

The corpus-level variational posterior for $\boldsymbol{\beta'}$ 
with truncation level $T$ is $q(\boldsymbol{\beta'}) = \prod_{i=1}^{T-1} \text{Beta}(\beta'_i | u_i, v_i)$, where $\{u_i, v_i\}_{i=1}^{T-1}$ are the corpus-level variational parameters. 

The ELBO of the training dataset can be written as:
\begin{eqnarray}
\loss(\mathcal{D}|\Phi, \psi) &=& E_{q(\boldsymbol{\beta'})}[\log \frac{P(\boldsymbol{\beta'}|\gamma)}{q(\boldsymbol{\beta'}|\boldsymbol{u},\boldsymbol{v})}] + \sum_{j=1}^D \{  E_{q(\boldsymbol{\nu}^{(j)})}[\log\frac{P(\boldsymbol{\nu}^{(j)}|\alpha)}{q(\boldsymbol{\nu}^{(j)})}] + \nonumber\\
& &\sum_{k=1}^{K} E_{q(\boldsymbol{\beta'}) q(c_{k}^{(j)}|\boldsymbol{\varphi}_{k}^{(j)})}[\log\frac{P(c_{k}^{(j)}|\boldsymbol{\beta})}{q(c_{k}^{(j)}|\boldsymbol{\varphi}_{k}^{(j)})}] +  E_{q(\boldsymbol{\nu}^{(j)})q( \boldsymbol{c}^{(j)})}[P(x^{(j)}|\boldsymbol{\nu}^{(j)}, \boldsymbol{c}^{(j)}, \Phi)]\}\nonumber\\
\end{eqnarray}
where 
$\boldsymbol{\beta} = \{\beta_i\}_{i=1}^{T}$, 
$\boldsymbol{\nu}^{(j)} = \{\nu_{k}^{(j)}\}_{k=1}^{K-1}$, $\boldsymbol{c}^{(j)} = \{c_{k}^{(j)}\}_{k=1}^K$, $\boldsymbol{\varphi}_{k}^{(j)} = \{\varphi_{ki}^{(j)}\}_{i=1}^{T}$. The details of 
 the derivation of the ELBO can be found in Appendix 1.5.

Gumbel-Softmax estimator~\citep{Jang2017Categorical} is used for backpropagating through the categorical random variables $\boldsymbol{c}$. Instead of joint training with the NN parameters, mean-field updates are used to learn the corpus-level variational parameters $\{u_i, v_i\}_{i=1}^{T-1}$:
\begin{eqnarray}
u_i = 1 + \sum_{j=1}^D \sum_{k=1}^K \varphi_{ki}^{(j)}; \quad v_i = \gamma + \sum_{j=1}^D \sum_{k=1}^K \sum_{l=i+1}^T \varphi_{kl}^{(j)}
\label{eqn:hie_update}
\end{eqnarray}

\subsection{Discussion}
\label{sec:discuss}

In \itmvae, we get the document-specific topic distribution $G^{(j)}$ by sampling the atom weights from a GEM. Instead of being drawn from a continuous base distribution, the atoms are modeled as trainable parameters as in~\cite{blei2003lda,srivastava2017avitm,miao2017discovering}. Thus, the atoms are shared by all documents naturally without
the need to use a hierarchical construction like HDP~\cite{Teh2006hdp}. The hierarchical extension, \itmvaehie, which models $G^{(j)}$ in a hierarchical manner, is more flexible and can generate topic distributions with better variability.
A detailed comparison is illustrated in Section~\ref{sec:itmvae-hie}.

\section{Experiments}
\label{sec:experiment}

In this section, we evaluate the performance of \itmvae and its variants on two public benchmarks: 20News and RCV1-V2, and demonstrate the advantage brought by the variants of \itmvae.
To make a fair comparison, we use exactly the same data and vocabulary 
as \citep{srivastava2017avitm}.



The configuration of the experiments is as follows. We use a two-layer fully-connected neural network for $g({\bf w}_{1:N};\psi)$
of Equation~\ref{eqn:kumar_parameter}, and the number of hidden units is set to $256$ and $512$ for 20News and RCV1-V2, respectively. 
The truncation level $K$
in Equation~\ref{eqn:inf_pi} is set to $200$ so that the maximum topic numbers will never exceed the ones used by baselines.
\footnote{In these
baselines, at most $200$ topics are used. Please refer to  Table~\ref{table:atc} for details.} The concentration parameter $\alpha$ for GEM distribution is
cross-validated on validation set from $[10,20,30,50]$ for \itmvae and \itmvaeprod.
Batch-Renormalization~\citep{Ioffe2017Batch}
is used to stabilize the training procedure. Adam~\citep{kingma2014adam} is used to optimize the model and 
the learning rate is set to $0.01$.
The code of \itmvae and its variants is available at \url{http://anonymous}.

\subsection{Perplexity and Topic Coherence}
\label{sec:perplexity_coh}
Perplexity is widely used by topic models to measure the goodness-to-fit capability,
which is defined as: $\exp (-\frac{1}{D}\sum_{j=1}^D\frac{1}{|{\bf x}^{(j)}|}\log p({\bf x}^{(j)}))$, where $D$ is the number of documents
, and $|{\bf x}^{(j)}|$ is the number of words in the $j$-th document ${\bf x}^{(j)}$. Following previous work,
the variational lower bound is used to estimate the perplexity.

\begin{table*}[ht]
\caption{Comparison of perplexity (lower is better) and topic coherence (higher is better) between different topic models on 20News and RCV1-V2 datasets.}
\begin{center}
\makebox[\textwidth][c]{
\begin{tabular}{l|cccccccc}
\hline
\multirow{2}{*}{Methods} & \multicolumn{4}{c}{Perplexity} &\multicolumn{4}{c}{Coherence}\\
\cmidrule(lr){2-5}\cmidrule(lr){6-9}& \multicolumn{2}{c}{20News} &\multicolumn{2}{c}{RCV1-V2}& \multicolumn{2}{c}{20News} &\multicolumn{2}{c}{RCV1-V2}\\
\cmidrule(lr){1-1}\cmidrule(lr){2-5}\cmidrule(lr){6-9}\#Topics
&$50$&$200$&$50$&$200$ &$50$&$200$&$50$&$200$\\
 \hline
LDA~\citep{hoffman2010online}$^\dagger$ & $893$ & $1015$ &$1062$ & $1058$ & $0.131$ & $0.112$ &\multicolumn{2}{c}{$-$}\\
DocNADE & $797$ & $804$ & $856$ & $670$ & $0.086$ & $0.082$ & $0.079$ & $0.065$\\
HDP~\citep{wang2011online}$^\dagger$ & \multicolumn{2}{c}{$937$} &\multicolumn{2}{c}{$918$} & \multicolumn{2}{c}{$-$} & \multicolumn{2}{c}{$-$}\\
NVDM$^\dagger$ & $837$ & $873$ &$717$ & $588$ & $0.186$ & $0.157$ &\multicolumn{2}{c}{$-$} \\
NVLDA & $1078$ & $993$ &$791$ & $797$ & $0.162$ & $0.133$ &$0.153$ & $0.172$ \\
ProdLDA & $1009$ & $989$ &$780$ & $788$ & $0.236$ & $0.217$ &$0.252$ & $0.179$\\
GSM$^\dagger$ &$787$ & $829$ &$653$ & $521$ &$0.223$ & $0.186$ &\multicolumn{2}{c}{$-$}\\
GSB$^\dagger$ &$816$ & $815$ &$712$ & $544$ &$0.217$ & $0.171$ &\multicolumn{2}{c}{$-$}\\
RSB$^\dagger$ &$785$ & $792$ &$662$ & $534$ &$0.224$ & $0.177$ &\multicolumn{2}{c}{$-$}\\
RSB-TF$^\dagger$ & \multicolumn{2}{c}{$788$} &\multicolumn{2}{c}{$532$}  & \multicolumn{2}{c}{$-$} & \multicolumn{2}{c}{$-$}\\
\cmidrule(lr){1-1}\cmidrule(lr){2-5}\cmidrule(lr){6-9}
\itmvae& \multicolumn{2}{c}{$877$} &\multicolumn{2}{c}{$1124$} & \multicolumn{2}{c}{$0.205$} &\multicolumn{2}{c}{$0.218$}\\
\itmvaeprod& \multicolumn{2}{c}{$\bf 775$} &\multicolumn{2}{c}{$\bf 508$} & \multicolumn{2}{c}{$0.278$} &\multicolumn{2}{c}{$0.3$}\\
\itmvaefull& \multicolumn{2}{c}{876} &\multicolumn{2}{c}{692} & \multicolumn{2}{c}{$0.285$} &\multicolumn{2}{c}{$\bf 0.311$}\\
\itmvaehie& \multicolumn{2}{c}{912} &\multicolumn{2}{c}{747} & \multicolumn{2}{c}{$\bf 0.29$} &\multicolumn{2}{c}{$0.27$}\\
\hline
\end{tabular}
}
\begin{minipage}{0.95\textwidth}
{\small
$\dagger$: 
We take these results from \citep{miao2017discovering} directly, since we use exactly the same datasets. The symbol "$-$" indicates that 
\citep{miao2017discovering} does not provide the corresponding values. 
As this paper is based on variational inference, we do not compare with LDA and HDP using Gibbs sampling, which are usually time consuming.
}

\end{minipage}
\end{center}
\label{table:atc}
\end{table*}
 As the quality of the learned topics is not directly reflected by perplexity~\citep{newman2010automatic},
topic coherence is designed to match the human judgment. 
We adopt 
NPMI~\citep{lau2014machine} as the measurement of topic coherence, as is adopted by \citep{miao2017discovering,srivastava2017avitm}.\footnote{We use the code provided by \citep{lau2014machine} at \url{https://github.com/jhlau/topic_interpretability/}}
We define a topic to be an {\it Effective Topic} if it 
becomes the top-$1$ significant topic of a sample among the training set more than $\tau \times D$ times, where $D$ 
is the training set size and $\tau$ is a ratio.
We set $\tau$ to $0.5\%$ in our experiments.
Following \citep{miao2017discovering}, we use an average over topic coherence
computed by top-5 and top-10 words 
across five random runs, which is more 
robust~\citep{lau2016sensitivity}. 

Table~\ref{table:atc} shows the perplexity and topic coherence of different topic models on 20News and RCV1-V2 datasets. 
We can clearly see that our models outperform the baselines, which indicates that our models have better goodness-to-fit capability and can discover topics that match more closely to human judgment. 
We can also see that \itmvaehie achieves better perplexity than \cite{wang2011online}, in which a similar hierarchical construction is used. 
Note that comparing the ELBO-estimated perplexity of \itmvaehie with other models directly is not suitable, as it has a lot more random variables, 
which usually leads to a higher ELBO. 
The possible reasons for the good coherence achieved by our models are  
1) The ``Product-of-Experts'' enables the model to model sharper distributions. 2) The 
nonparametric characteristic means the models can adapt the number of topics to data, thus topics can be
sufficiently trained and of high diversity. Table 1 in Appendix 1.1 illustrates the topics learned by \itmvaeprod. Please refer to Appendix 1.1 in the supplementary for more details.

\subsection{The Effect of Hyper-Prior on \itmvae}
\label{sec:itmvae_bnprior}

In this section, we provide quantitative evaluations on the effect of the hyper-prior for \itmvae.
Specifically, a relatively non-informative hyper-prior $\text{Gamma}(1,0.05)$
 is imposed on $\alpha$. 
And we initialize the global variational parameters $\gamma_1$ and
$\gamma_2$ of Equation~\ref{eqn:itmvae_elbo_fb} the same as the non-informative Gamma prior. 
Thus the expectation of $\alpha$ given the variational posterior $q(\alpha|\gamma_1,\gamma_2)$ is $20$ 
before training.
A SGD optimizer with a learning rate of $0.01$ is used to optimize $\gamma_1$ and $\gamma_2$. 
No KL annealing and decoder regularization are used for \itmvaefull.
\begin{mixedcol}
\begin{multicols}{2}
    \begin{figure}[H]
    \begin{center}
    \includegraphics[width=0.75\linewidth]{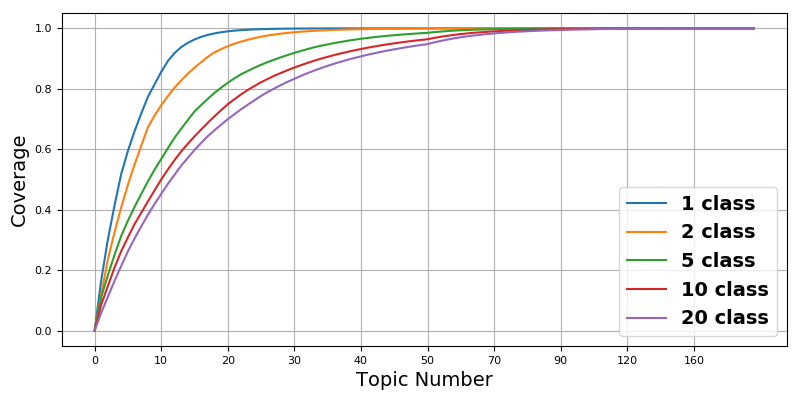}\label{fig:assignment_cdf}
    \end{center}
    \caption{Topic coverage w.r.t number of used topics learned by \itmvaefull.}
    \end{figure}
	
    \begin{table}[H]
        \caption{The posterior distribution of $\alpha$ learned by \itmvaefull on subsets of 20News dataset.}
    \begin{center}
    \begin{tabular}{ccccc}
    \hline
    \#classes & $\gamma_1$ & $\gamma_2$ & $\expect_{q(\alpha)}[\alpha]$ \\ 
     \hline
    1 & $16.88$ & $4.58$ & $3.68$\\
    2 & $23.03$ & $3.68$ & $6.25$\\
    5 & $31.43$ & $2.88$ & $10.93$\\
    10& $39.64$ & $2.69$ & $14.71$\\
    20& $48.91$ & $2.98$ & $16.39$\\
    \hline
    \end{tabular}
    \end{center}
    \label{table:prod_alpha}
    \end{table}
\end{multicols}
\vspace{-5mm}
\end{mixedcol}

Table~\ref{table:prod_alpha} reports 
the learned global variational parameter $\gamma_1$, $\gamma_2$ and the expectation of $\alpha$ given the variational
poster $q(\alpha|\gamma_1,\gamma_2)$ on several subsets of 20News dataset, which contain $1$, $2$, $5$, $10$ and $20$ classes, respectively.\footnote{Since there are no labels for the 20News dataset provided by \citep{srivastava2017avitm}, we preprocess the dataset ourselves in this illustrative experiment.}
We can see that, once the training is done, the variational posterior  $q(\alpha|\gamma_1,\gamma_2)$ is very confident, and $\expect_{q(\alpha|\gamma_1,\gamma_2)}[\alpha]$, the expectation of $\alpha$ given the variational posterior, is adjusted to the training set. 
For example, if the training set
contains only $1$ class of documents, $\expect_{q(\alpha|\gamma_1,\gamma_2)}[\alpha]$ after training is $3.68$, 
Whereas, when the training set consists of $10$ classes of documents, 
$\expect_{q(\alpha|\gamma_1,\gamma_2)}[\alpha]$ after training is $14.71$.
This indicates that \itmvaefull can learn to adjust $\alpha$ to
data, thus the number of discovered topics will adapt to data better. 
In contrast, for \itmvaeprod (without the hyper-prior), when the decoder is strong, no matter how many classes the dataset contains, 
the number of topics will be constrained tightly due to
the {\it collapse-to-prior} problem of AEVB,
and
KL-annealing and decoder regularizing tricks do not help much.

Figure~\ref{fig:assignment_cdf} illustrates the {\it training set coverage} w.r.t the number of used topics when the training set
contains $1$, $2$, $5$, $10$ and $20$ classes, respectively. Specifically, we compute the average weight of every topic on the training dataset, and sort the topics according to their average weights. The topic coverage is then defined as the cumulative sum of these weights. 
Figure~\ref{fig:assignment_cdf}
shows that, with the increasing of the number of classes, more topics are utilized by \itmvaefull to reach the same
level of topic coverage, which indicates that the model has the ability to adapt to data.

\subsection{The Evaluation of \itmvaehie}
\label{sec:itmvae-hie}
In this section, by comparing the topic coverage and sparsity\footnote{To compare the sparsity of the posterior topic proportions of each model, we sort the topic weights of every training document and average across the dataset. Then, the logarithm of the average weights are plotted w.r.t the topic index.} of \itmvaeprod and \itmvaehie, we show that the hierarchical construction can help the model to learn more topics, and produce posterior topic proportions with higher sparsity.

The model configurations are the same for \itmvaeprod and \itmvaehie, except that $\alpha$ is set to $5$ and $20$ for \itmvaeprod, and $\gamma=20, \alpha=5$ for \itmvaehie. For \itmvaehie, the corpus-level updates are done every $200$ epochs on 20News, and $20$ epochs on RCV1-V2.

\begin{figure*}[ht]
\begin{center}
\subfigure[Coverage]{\includegraphics[width = 0.4\linewidth]{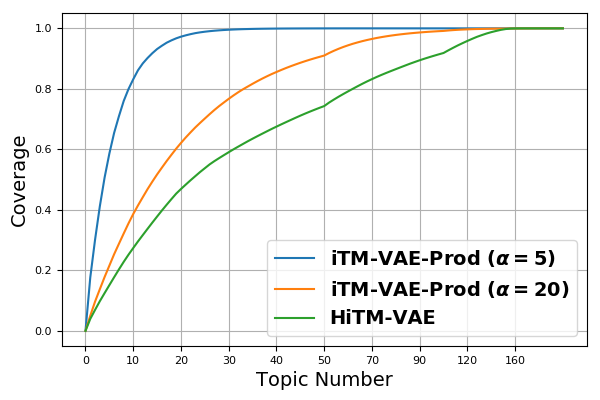}\label{fig:hie_coverage}}
\hspace{0.05\linewidth}
\subfigure[Sparsity]{\includegraphics[width = 0.4\linewidth]{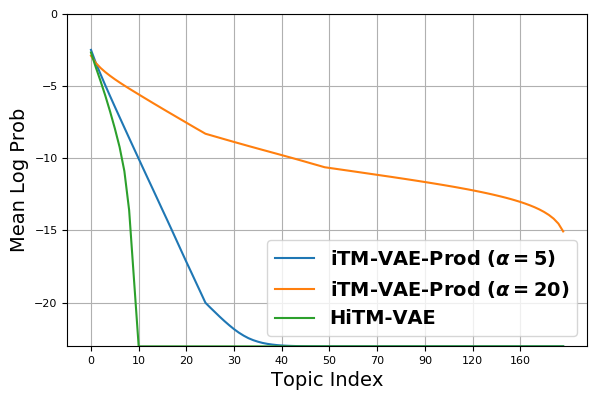}\label{fig:hie_sparsity}}
\end{center}
   \caption{Comparison of the topic coverage~(a) and sparsity~(b) between \itmvaeprod ($\alpha=5$), \itmvaeprod ($\alpha=20$) and \itmvaehie ($\gamma=20$, $\alpha=5$). We can see that \itmvaehie can simultaneously discover more topics and produce sparser posterior topic proportions. 
   }\label{fig:hie_cs}
\end{figure*}

As shown in Figure~\ref{fig:hie_cs}, \itmvaehie can learn more topics than \itmvaeprod ($\alpha=20$), and the sparsity of its posterior topic proportions is significantly higher. \itmvaeprod ($\alpha=5$) has higher sparsity than \itmvaeprod ($\alpha=20$). However, its sparsity is still lower than \itmvaehie with the same document-level concentration parameter $\alpha$, and it can only learn a small number of topics, which means that there might exist rare topics that are not learned by the model. The comparison of \itmvaehie and \itmvaeprod ($\alpha=5$) shows that the superior sparsity not only comes from a smaller per-document concentration hyper-parameter $\alpha$, but also from the hierarchical construction itself. 



\section{Conclusion}
\label{sec:conclusion}


In this paper, we propose \itmvae and \itmvaeprod, which are nonparametric topic models that are modeled by Variational Auto-Encoders.  Specifically, a stick-breaking prior is used to generate
the atom weights of countably infinite shared topics, and the \kumar distribution is
exploited such that the model can be optimized by AEVB algorithm. We also propose \itmvaefull which introduces a hyper-prior into the VAE framework such that the model can adapt better to data. This technique is general and can be incorporated into other VAE-based models to alleviate the {\it collapse-to-prior} problem. To further diversify the document-specific topic distributions, we use a hierarchical construction in the generative procedure. And we show that the proposed model \itmvaehie can learn more topics and produce sparser posterior topic proportions. The advantage of \itmvae and its variants over traditional nonparametric topic models is that the inference is performed by feed-forward neural networks, which is of rich representation capacity and requires only limited knowledge of the data. Hence, it is flexible to incorporate 
more information sources to the model, and we leave it to future work.
Experimental results on two public benchmarks show that \itmvae and its variants
outperform the state-of-the-art baselines.

\bibliographystyle{plainnat}
\bibliography{nips_2018}

\end{document}